\documentclass[runningheads]{llncs}
\usepackage[title]{appendix}
\usepackage{graphicx}
\usepackage{subcaption}
\usepackage[linesnumbered,ruled,vlined]{algorithm2e}
\usepackage{float}
\usepackage{hyperref}
\usepackage{amsmath,amssymb}
\usepackage{math_definition}
\usepackage[table,xcdraw,dvipsnames]{xcolor}
\usepackage{cite}
\usepackage{color}
\usepackage{multirow}
\usepackage{comment}
\usepackage{tabularx}
\usepackage{booktabs}
\usepackage{listings}
\usepackage[misc]{ifsym} 
\usepackage{orcidlink}
\newcommand\blfootnote[1]{
\begingroup
\hypersetup{hidelinks}
\renewcommand\thefootnote{}\footnote{#1}
\addtocounter{footnote}{-1}
\endgroup
}

\newcommand{\reviewflag}[1]{\textcolor{black}{#1}}

\newcommand{\planchange}[1]{\textcolor{black}{#1}}

\newcommand{\sotaupdate}[1]{\textcolor{black}{#1}}
\newcommand{\baselineupdate}[1]{\textcolor{black}{#1}}

\allowdisplaybreaks
\newcommand{\benchsize}[2]{\shortstack{#1\\(#2)}}
\newcommand{\papertablefont}{\fontsize{8pt}{10pt}\selectfont}
\begin{document}
\title{A Hybrid SMT-NRA Solver: Integrating 2D-Cell-Jump-Based Local Search, MCSAT and OpenCAD}

\titlerunning{A Hybrid SMT-NRA Solver}

\author{%
   Tianyi Ding\inst{1}$^\textrm{(\Letter)}$\orcidlink{0009-0009-5309-3259} \and
   Haokun Li\inst{1}$^\textrm{(\Letter)}$\orcidlink{0000-0001-6411-9324}\and
   Xinpeng Ni\inst{1}$^\textrm{(\Letter)}$\orcidlink{0009-0005-8967-5665}\and
   Bican Xia\inst{1}$^\textrm{(\Letter)}$\orcidlink{0000-0002-2570-2338} \and
   Tianqi Zhao\inst{2}$^\textrm{(\Letter)}$\orcidlink{0000-0001-5099-0805}
}

 \authorrunning{T. Ding et al.}

\institute{
    School of Mathematical Sciences, Peking University, Beijing, China\\
    \email{\{dingty,nxp\}@stu.pku.edu.cn}, 
    \email{xbc@math.pku.edu.cn}
    \email{ker@pm.me}
    \and
    Zhongguancun Laboratory, Beijing, China\\
    \email{zhaotq@zgclab.edu.cn}
}
%

%
\maketitle              

\blfootnote{The authors make equal contribution to this work and are listed in alphabetical order.}

\vspace{-1cm}
\begin{abstract}
In this paper, we propose a hybrid framework for Satisfiability Modulo the Theory of Nonlinear Real Arithmetic (SMT-NRA for short). First, we introduce a two-dimensional cell-jump move, called \emph{2D-cell-jump}, generalizing the key operation, cell-jump, of the local search method for SMT-NRA. In contrast to the existing cell-jump, which updates assignments exclusively along a given line, the 2D-cell-jump enables search over an arbitrarily selected plane. Then, we propose an extended local search framework, named \emph{2D-LS} (following the existing local search framework, LS, for SMT-NRA). Finally, we present a hybrid framework for SMT-NRA integrating 2D-LS, MCSAT and OpenCAD, to improve search efficiency through information exchange. The experimental results demonstrate improvements in local search performance, highlighting the effectiveness of the proposed methods.
\end{abstract}
%
%
\section{Introduction}
\label{sec:intro}
Satisfiability Modulo Theories (SMT) is concerned with determining the satisfiability of first-order logic formulas under background theories, such as integer arithmetic, real arithmetic, arrays, bit vectors, strings, and others. This paper concentrates on SMT problems over the theory of {quantifier-free} nonlinear real arithmetic (NRA), referred to as SMT-NRA. The goal is to determine the satisfiability of {\em polynomial formulas}, which are expressed in the form of $\bigwedge _{i} \bigvee_{j} p_{ij}(\vX) \rhd_{ij} 0, \text{where} ~\rhd_{ij}\in \{<, >,\leq,\geq,=,\neq\}$ and $p_{ij}(\vX)$ are polynomials.
SMT-NRA has found widespread applications in fields where nonlinear constraints are common, including verification of hybrid systems \cite{Alur2011FormalVO,cimatti2013smt,chen2018control}, software/hardware verification \cite{hardVeri2016,beyer2018unifying,li2010scalable,faure2024methodology}, automated theorem proving \cite{weber2011smt,leino2013automating,keller2019smtcoq}, and cryptology \cite{akinyele2013using,gundaram2021performance,baelde2025smt}.


Tarski proposed an algorithm in 1951 \cite{tarski1998decision}, solving the problem of quantifier elimination (QE) of the first-order theory over real closed fields, which takes SMT-NRA as a special case. Cylindrical Algebraic Decomposition (CAD), another real QE method introduced by Collins in 1975 \cite{collins1975quantifier}, can solve polynomial constraints by decomposing space into finitely many regions (called {\em cells}) arranged cylindrically. CAD provides a more practical approach to quantifier elimination than Tarski's procedure though it remains of doubly exponential complexity.
In practice, the Model-Constructing Satisfiability Calculus (MCSAT) \cite{mcsat.Moura.2013} is  a widely used complete SMT algorithms. MCSAT integrates two solvers from the classical framework into one solver that simultaneously searches for models in both the Boolean structure and the theory structure, thereby constructing consistent Boolean assignments and theory assignments. 
Several state-of-the-art (SOTA) SMT solvers supporting NRA have been developed over the past two decades. Representative SOTA solvers include Z3 \cite{z32008} and Yices2 \cite{Yices2}, which implement MCSAT, as well as CVC5 \cite{cvc52022} and MathSAT5 \cite{mathSAT5}, which use alternative techniques.

The computational complexity of SMT-NRA remains a challenge, motivating research of incomplete solvers that are usually more efficient in finding SAT assignments. Local search,  a popular paradigm,
has been developed in recent years for SMT-NRA \cite{ls2023,realmulcaiLS,realcaiLS}. 
Local search begins with a theory assignment and approaches a model of the polynomial formula iteratively by moving locally. This process ends when a model is found or other termination conditions are met. The most effective move for real-space search is the `cell-jump' proposed by Li et al. \cite{ls2023}, which leads sample points to different CAD cells via one-dimensional moves. 


The complementarity between complete methods and local search has led to the development of hybrid solvers for related problems, suggesting a similar hybrid approach may hold promise for solving SMT-NRA. Notable examples include the hybridization of Conflict-Driven Clause Learning (CDCL) and local search for SAT \cite{hybSAT2008}, the combination of CDCL(T) and local search for SMT-NIA \cite{deep2021}, and the integration of local search with MCSat for SMT-NIA \cite{10.1007/978-3-031-99984-0_6}. While the LS-MCSat framework proposed in \cite{10.1007/978-3-031-99984-0_6} is theory-agnostic in its architecture, the concrete move operations must be implemented by each theory plugin according to its specific characteristics. For instance, the NIA plugin leverages its maintained feasible intervals to implement a lightweight ``feasible-sets jumping," whereas an NRA plugin typically relies on polynomial root isolation to perform ``cell-jumping."

To efficiently handle unsatisfiable formulas dominated by algebraic conflicts,
we integrate OpenCAD \cite{han2014constructing} into our hybrid framework.
OpenCAD is a lightweight variant of CAD
designed specifically for formulas consisting solely of strict inequalities.
Unlike CAD, which constructs cells of all dimensions,
OpenCAD computes only \emph{full-dimensional cells} (regions with non-zero
volume), reducing computational overhead.

In this paper, we aim at advancing local search in SMT-NRA on problems in the form of \begin{equation*}
\bigwedge _{i} \bigvee_{j} p_{ij}(x_1,\ldots,x_n) \rhd_{ij} 0, \text{where} ~\rhd_{ij}\in \{<, >,\neq\} ~\text{and}~ p_{ij}\in \mathbb{Q}[x_1,\ldots,x_n],
\end{equation*}
and make the following contributions:

\begin{itemize}
    \item 
    We propose a novel cell-jump mechanism, termed 2D-cell-jump, which enables effective next moves even when the local search gets trapped in a local optimum. 
    \item We propose an extended local search framework, named {2D-LS}, integrating the MCSAT framework to improve search efficiency. 
    \item We design a hybrid framework for SMT-NRA that exploits the complementary strengths of  MCSAT, 2D-LS and OpenCAD. In this framework, 2D-LS accelerates model search within MCSAT and assists in identifying unsatisfiable cells, which are then used as a heuristic signal to trigger a switch to OpenCAD for handling unsatisfiable formulas dominated by algebraic conflicts.
    \item The above proposed methods have been implemented as a solver called HELMS. Comparison to SOTA solvers on a large number of benchmarks shows that the newly proposed methods are effective. 
\end{itemize}


The rest of this paper is organized as follows. Sect. \ref{sec:pre} introduces the preliminaries of SMT-NRA and local search techniques for this problem. Sect. \ref{sec:ls-expand} extends traditional local search to a 2D-cell-jump-based variant. Sect. \ref{sec:hybrid} outlines the hybrid framework that integrates 2D-cell-jump local search, MCSAT, and OpenCAD. Finally, Section \ref{sec:exp} presents experimental results, demonstrating that our hybrid SMT-NRA solver is efficient and effectively leverages the complementary strengths of 2D-LS, MCSAT, and OpenCAD. 

\section{Preliminaries}
\label{sec:pre}

\subsection{Problem Statement}
\label{subsec:pre-SMT-NRA}
Let $\vX=(x_1,\ldots,x_n)$ be a vector of variables.
Denote by $\Q$, $\R$, $\Z$ and $\N$ the set of rational numbers, real numbers, integers and natural numbers, respectively. 
Let $\Q[\vX]$ be the ring of polynomials in the variables $x_1,\ldots,x_n$ with coefficients in $\Q$.

For any \(1 \le i \le n\), let \(e_i = (0, \ldots, 1, \ldots, 0) \in \mathbb{R}^n\) denote the \(i\)-th unit vector. For any point \(\alpha \in \mathbb{R}^n\) and vector \(d \in \mathbb{R}^n\), define the line \(\alpha + \langle d \rangle := \{ \alpha + \lambda d \mid \lambda \in \mathbb{R} \}\). Similarly, for linearly independent \(d_1, d_2 \in \mathbb{R}^n\), define the plane \(\alpha + \langle d_1, d_2 \rangle := \{ \alpha + \lambda_1 d_1 + \lambda_2 d_2 \mid \lambda_1, \lambda_2 \in \mathbb{R} \}\). In particular, \((0,\ldots,0) + \langle e_i, e_j \rangle\) is an \emph{axes plane}, and \(\alpha + \langle e_i, e_j \rangle\) is a plane parallel to an axes plane.

\begin{definition}[Polynomial Formula]
    The following formula 
    \begin{equation}\label{eq:problem}
        F=\bigwedge _{i=1}^M \bigvee_{j=1}^{m_i} p_{ij}(\vX) \rhd_{ij} 0
    \end{equation}
    is called a \emph{polynomial formula}, where $1\leq i\leq M<+\infty, 1\leq j\leq m_i<+\infty$, $p_{ij}\in\Q[\vX]$ and $\rhd_{ij} \in\{<,>,\leq,\geq,=,\neq\}$. Moreover, 
    $\bigvee_{j=1}^{m_i} p_{ij}(\vX) \rhd_{ij} 0$ is called a \emph{polynomial clause} or simply a \emph{clause}, and $p_{ij}(\vX) \rhd_{ij} 0$ is called an \emph{atomic polynomial formula} or simply an \emph{atom}. 
    As is customary, we call an atomic polynomial formula or its negation a \emph{polynomial literal} or simply a \emph{literal}.
\end{definition}

For any polynomial formula $F$, a \emph{complete assignment} is a point $\alpha=(a_1,\ldots,\allowbreak a_n)\allowbreak\in\R^{n}$. An atom is \emph{true} under $\alpha$ if it evaluates to true under $\alpha$, and otherwise it is \emph{false} under $\alpha$,
A clause is \emph{satisfied} under $\alpha$ if
at least one atom in the clause is true under $\alpha$, and \emph{falsified} under $\alpha$ otherwise. 
A polynomial formula is \emph{satisfiable} {\rm(}SAT{\rm)} if there exists a complete assignment in $\R^n$ such that all clauses in the formula are satisfied, and such an assignment is a \emph{model} to the polynomial formula. 
A polynomial formula is \emph{unsatisfiable} {{\rm(}UNSAT{\rm)}} if any assignment is not a model.

The problem addressed in this paper is the satisfiability of polynomial formulas of the form (\ref{eq:problem}), where $\rhd_{ij} \in\{<,>,\neq\}$. This restricted choice of relations is motivated by practical implementation constraints: the base solvers employed in our prototype currently support only strict inequalities. Handling nonlinear equations would require sophisticated symbolic algebra implementation (see for instance 
\cite{XIA2002633} and \cite{chen2025geometry}), which are not yet integrated into the current version. Alternative approaches, such as that in \cite{realcaiLS}, circumvent this implementation difficulty by relaxing equations and non-strict inequalities with a small perturbation parameter $\varepsilon$ (\eg relaxing $p=0$ to $p>-\varepsilon \land p<\varepsilon$). 

\subsection{Local Search for SMT-NRA}
\label{subsec:pre-ls}

Local search for SMT-NRA starts from an initial assignment  $(x_1,\ldots,x_n)\mapsto(a_1^{(0)},\ldots,a_n^{(0)})$. At each step, the algorithm performs a move that updates the assignment until either the assignment satisfies the given formula $F$ or another terminal condition is reached. The core of local search is its move operator, and for SMT-NRA this move operator is the cell-jump operation \cite[Alg. 3]{ls2023}. This operation updates the assignment to flip the sign of a certain polynomial in $F$, attempting to satisfy more constraints in the formula. Scoring function is to assess candidate moves.

\textbf{Scoring function.}
For a literal \(\ell: p \rhd 0\) (\(\rhd \in \{>, <\}\)), let
\[
L2O(\ell, \alpha) =
\begin{cases}
0, & \text{if } \ell \text{ is true under } \alpha,\\
|p(\alpha)| + \delta, & \text{otherwise},
\end{cases}
\]
with \(\delta > 0\). For a clause \(c\), \(L2O(c, \alpha) = \min_{\ell \in c} L2O(\ell, \alpha)\); for a CNF formula \(F\), \(L2O(F, \alpha) = \sum_{c \in F} L2O(c, \alpha) \cdot w(c)\), where $w(c)$ is the dynamic weight of clause $c$. 
The score of a move \(\alpha \to \alpha'\) is \(score(\alpha, \alpha') = L2O(F, \alpha) - L2O(F, \alpha')\). 
It follows that \(L2O(F, \alpha) = 0\) if and only if \(\alpha\) satisfies \(F\); otherwise, the value is positive. A move has a positive score if it decreases the L2O value, indicating that the new assignment is closer to satisfying the formula; a higher score corresponds to a greater improvement.

From a geometric perspective, an assignment is a point in \(\mathbb{R}^n\), and the set of polynomials in \(F\) decomposes \(\mathbb{R}^n\) into finitely many cells, with the current assignment lying in exactly one of them. Flipping the sign of a polynomial therefore corresponds to jumping to another cell. The cell-jump operation seeks such a transition by moving along a line in \(\mathbb{R}^n\). Two types of cell-jump have been proposed, depending on the chosen direction of this line: (1) a cell-jump along a coordinate axis direction, which changes only the assignment of the corresponding variable; and (2) a cell-jump along a given arbitrary direction, which may change the assignments of all variables. From an algebraic perspective, once the direction is fixed, the polynomial to be flipped becomes a univariate polynomial, and sign change is detected via real-root isolation.

\section{2D-Cell-Jump-Based Local Search}
\label{sec:ls-expand}

In Sect. \ref{subsec:sample}, we propose a new cell-jump operation, called \emph{2D-cell-jump}. 
Comparing to \emph{cell-jump} in \cite[Alg. 2]{ls2023}, 2D-cell-jump allows searching for updating assignments in a 2D plane instead of a line. 
In Sect. \ref{subsec:els}, based on 2D-cell-jump, we develop a new local search algorithm for SMT-NRA, called 2D-LS. 
The algorithm can be considered as an extension of LS \cite[Alg. 3]{ls2023}.





\subsection{2D-Cell-Jump}
\label{subsec:sample}

Remark that the cell-jump operation proposed in \cite{{ls2023}} is limited to searching for a solution along a straight line, which is one-dimensional. 
With the assistance of MCSAT, the search process can be extended into higher dimensional space. 
In this subsection, we define $2$-dimensional cell-jump, 2D-cell-jump for short, expanding the cell-jump move from a line parallel to a coordinate axis to a plane parallel to an axes plane, and from a given line to a given plane. 
Theoretically, this approach can be generalized to $d$-dimensional space, where $d\ge2$. Given that local search is a lightweight algorithm, each assignment update should incur only low-cost computation. In our approach, however, each local search update requires invoking the MCSAT procedure on $d$ variables. Fortunately, when $d=2$, the MCSAT procedure is highly efficient, making it the most suitable choice for cell-jump operation.


Sample points are candidate assignments to move to in the cell-jump operation. 
To define the new cell-jump operation, we first introduce new sample points. 
\begin{definition}[Sample Point]
\label{def:sample-point}
Let \(\ell\) be a polynomial atom. Any assignment \(\alpha\in\R^n\) that satisfies \(\ell\) is called a \emph{sample point} of \(\ell\).
\end{definition}


We propose two variants of 2D-cell-jump, classified by the direction of the plane used for the move. In a plane parallel to an axes plane $\langle e_i,e_j\rangle$, the jump updates only the assignments of two variables $x_i$ and $x_j$; in other planes, it changes assignments of all variables simultaneously.
\begin{definition}[2D-Cell-Jump in a Plane Parallel to an Axes Plane]
Let $\ell$ be an atom that is false under the current assignment $\alpha$. If there exists a sample point $\alpha'$ of $\ell$ in $\alpha+\langle e_i,e_j\rangle$ for some $1\le i<j\le n$, then updating the assignment from $\alpha$ to $\alpha'$ is called a 2D-cell-jump in a plane parallel to an axes plane.
\end{definition}

\textbf{Generating sample points for 2D-Cell-Jump in a plane parallel to an axes plane.}
Suppose \(\ell: p(\vX) > 0\) (or \(< 0\)) is false under \(\alpha = (a_1, \ldots, a_n)\). For each pair \(x_i, x_j\) (\(i < j\)), we treat all other variables as fixed to their \(\alpha\)-values, reducing \(p\) to a bivariate polynomial \(p^{*}(x_i, x_j)\). If MCSAT returns a satisfying assignment \((b_i, b_j)\) for \(p^{*}(x_i, x_j) > 0\) (or \(< 0\)), then the assignment \(\alpha[x_i \mapsto b_i, x_j \mapsto b_j]\) is a sample point in the plane $\alpha+\langle e_i,e_j\rangle$.

\begin{definition}[2D-Cell-Jump in a Given Plane]
Let $\ell$ be an atom that is false under the current assignment $\alpha$, and let $\alpha+\langle d_1,d_2\rangle$ be a given plane. If there exists a sample point $\alpha'$ of $\ell$ in $\alpha+\langle d_1,d_2\rangle$, then updating the assignment from $\alpha$ to $\alpha'$ is called a 2D-cell-jump in the plane $\alpha+\langle d_1,d_2\rangle$.

\end{definition}

\textbf{Generating sample points for 2D-Cell-Jump in a given plane.}
Suppose \(\ell: p(\vX) > 0\) (or \(< 0\)) is false under \(\alpha = (a_1, \ldots, a_n)\). Given a plane $\alpha+\langle d_1,d_2\rangle$, we introduce two new variables $t_1,t_2$ and replace every $x_i$ with $a_i + d_{1,i} t_1 + d_{2,i} t_2$ in $p(\vX)$ to obtain a bivariate polynomial $p^{*}(t_1,t_2)$. If MCSAT returns a satisfying assignment \((c_1, c_2)\) for \(p^{*}(t_1, t_2) > 0\) (or \(< 0\)), then the assignment \((a_i + d_{1,i}c_1+d_{2,i} c_2)_{i=1}^n\) is a sample point in the plane $\alpha+\langle d_1,d_2\rangle$.

A difference between the 2D-cell-jump and the 1D-cell-jump lies in their search dimensions. If there exists a cell-jump along a line \(\alpha + \langle d_1 \rangle\) with a positive score, then for any vector \(d_2\) linearly independent of \(d_1\), there exists a 2D-cell-jump on the plane \(\alpha + \langle d_1, d_2 \rangle\) with a positive score, because the line is contained in the plane. The converse does not necessarily hold: a positive-score 2D-cell-jump on a plane does not imply the existence of a positive-score cell-jump on a particular line within that plane. This geometric inclusion does not guarantee that a witness returned by MCSAT for a particular atom and plane will have a positive score whenever an improving point exists. The larger search domain provides additional opportunities for search progression when the generated line-based moves offer no improvement.

\begin{example}\label{ex:line-plane-comparison}
Consider $F=\ell_1\land\ell_2$, where
\[
\ell_1:\ x^2+y^2-z^2<0,
\qquad
\ell_2:\ (x-3)^2+y^2+(z+1/2)^2-7<0.
\]
We compare one cell-jump with one 2D-cell-jump from $\alpha=(0,0,2)$, which satisfies only $\ell_1$ (Fig.~\ref{fig:line-plane-comparison}).
In the $x$-direction, the cell-jump move is confined to $y=0,z=2$, a line that misses the feasible region; the illustrated target $\beta=(3,0,2)$ satisfies only $\ell_2$.
The $xz$-plane ($y=0$) intersects the feasible region at points such as $\gamma=(3/2,0,8/5)$.
If the single sample point returned by MCSAT for $\ell_2$ in this plane also satisfies $\ell_1$, the 2D-cell-jump reaches a model of $F$.

\vspace{-5mm}
\begin{figure}
    \centering
    \begin{minipage}[b]{0.49\linewidth}
        \centering
        \includegraphics[width=\linewidth,trim=0 20bp 0 30bp,clip]{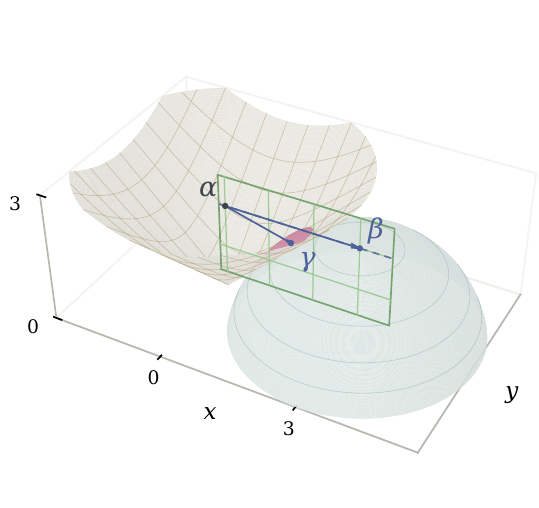}
        \par\small (a) Spatial view
    \end{minipage}\hfill
    \begin{minipage}[b]{0.49\linewidth}
        \centering
        \includegraphics[width=0.96\linewidth]{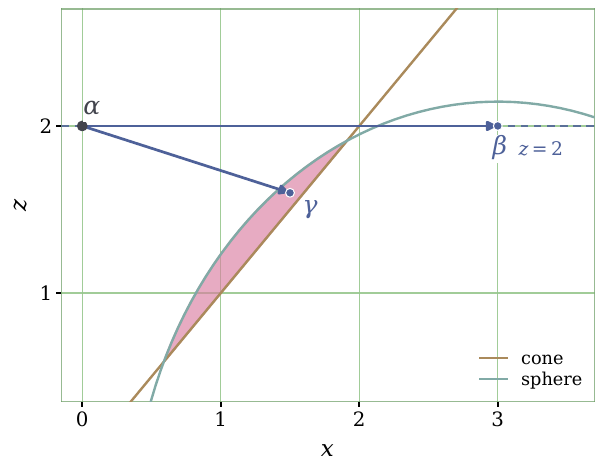}
        \par\small (b) Plane $y=0$
    \end{minipage}
    \caption{Comparsion of a cell-jump and a 2D-cell-jump.}
    \label{fig:line-plane-comparison}
\end{figure}
\end{example}

\vspace{-10mm}
\subsection{New Local Search Algorithm: 2D-LS}
\label{subsec:els}

Based on the new cell-jump operation, we develop a new local search algorithm, named 2D-LS (Alg.~\ref{alg:els}), which adopts a four-step search framework.

\begin{enumerate}
\item Try to find cell-jumps along all $n$ coordinate directions through the current assignment, and select one with the highest positive score.
\item If the first step fails, generate $2m$ random vectors $d_1,\ldots,d_{2m}$. Attempt to perform a cell-jump operation along a random line, which passes through the current assignment point with direction $d_i$.
\item If the second step fails, try to find 2D-cell-jumps in all $\binom{n}{2}$ planes through the current assignment parallel to coordinate planes, and select one with the highest positive score among the generated sample points.
\item If the third step fails, attempt to perform a 2D-cell-jump operation in a random plane, which is spanned by vectors $d_i$, $d_j$ and passes through the current assignment point.
\end{enumerate}

Local search can stagnate when none of its generated line-based moves has a positive score. The third and fourth steps expand the search to planes through the current assignment to help escape local optima of the line-based search. Geometrically, the solution set of a polynomial formula consists of connected regions in $\mathbb{R}^n$. LS searches for solutions along lines, while planar search offers more opportunities for intersection, since a plane can intersect a region missed by a search line contained in it. These steps dispense with the proximity preference used in the original cell-jump~\cite{ls2023}, since a useful next assignment may lie far from the current point. Candidate generation is heuristic: for each false atom and plane, MCSAT returns at most one satisfying assignment of the restricted atom without a distance constraint (Sect.~\ref{subsec:sample}). The positive-score tests and highest-score selection consider only these generated points.

\begin{algorithm}[t]
\scriptsize
\caption{2D-LS}  
\label{alg:els}

\KwIn{$F$, a polynomial formula with variables $x_1,\ldots,x_n$ such that the relational operator of every atom is `$<$' or `$>$'; $MaxRestart$, $MaxJump$ and $m$, three positive integers
}
\KwOut{$status$, `SAT' or `UNKNOWN'; $\alpha$, an assignment; $numJump$, a nonnegative integer}
    $numRestart\gets1$, $numJump\gets0$
    
    \While{$numRestart\leq MaxRestart$}{
        $\alpha\gets(a_1,\ldots,a_n)$, an initial complete assignment in $\Q^{n}$

        $k\gets0$
        
        \While{$k<MaxJump$}{
        \If{$F(\alpha)=$True}
        {\textbf{return} `SAT', $\alpha$, $numJump$}
        \uIf{a positive-score cell-jump exists along one of the $n$ coordinate directions\label{line:cp-1}
        }{
        perform such an operation with the highest score to update $\alpha$
        }

        \Else{
        generate $2m$ random vectors $d_1,\ldots,d_{2m}$, where $d_i\in\Q^{n}$
        
        $L\gets\{\alpha+\langle d_1\rangle ,\ldots,\alpha+\langle d_{2m}\rangle\}$

        \uIf{a positive-score cell-jump exists along a line in $L$}{
        perform such an operation with the highest score to update $\alpha$\label{line:cp-2}}

        
        \uElseIf{a positive-score 2D-cell-jump exists in one of the $\binom{n}{2}$ planes $\alpha+\langle e_i,e_j\rangle$ ($1\leq i<j\leq n$)\label{line:cp-3}}
        {perform such an operation with the highest score to update $\alpha$}
        \Else{
            $P\gets\{\alpha+\langle d_1,d_2\rangle,\ldots,\alpha+\langle d_{2m-1},d_{2m}\rangle\}$

            \uIf{a positive-score 2D-cell-jump exists in a plane in $P$}
            {perform such an operation with the highest score to update $\alpha$\label{line:cp-4}}
            \Else{\textbf{break}}
        }

        }

        $numJump$++; $k$++
        }
        \If{$F(\alpha)=$True}
        {\textbf{return} `SAT', $\alpha$, $numJump$}
        $numRestart$++
    }
    \textbf{return} `UNKNOWN', $\alpha$, $numJump$

\end{algorithm}

\section{A Hybrid Framework for SMT-NRA}
\label{sec:hybrid}

\begin{figure}[t]
    \centering
    \includegraphics[width=\linewidth]{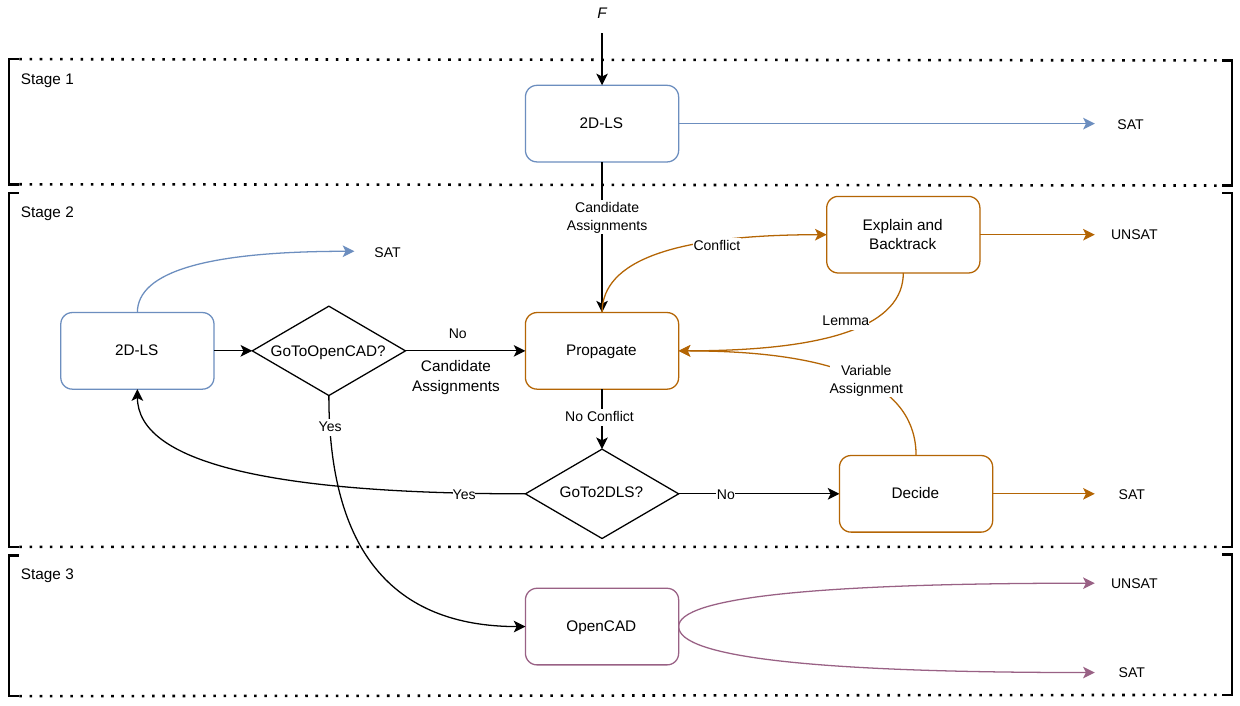}
    \caption{Overview of the workflow in the hybrid framework. The paths and modules shaded in blue, yellow, and purple denote the original components inherited from 2D-LS, MCSAT, and OpenCAD, respectively.}
    \label{fig:drawio}
    \vspace{-5mm}
\end{figure}

To make local search applicable to practical SMT-NRA solving, we further investigate its effective coupling with complete methods and propose a hybrid framework that integrates 2D-LS, MCSAT \cite{mcsat.Moura.2013} and OpenCAD \cite{han2014constructing}. 
We present the workflow overview of our hybrid framework in Fig. \ref{fig:drawio}, demonstrating the interaction mechanism among the three core modules—2D-LS, MCSAT, and OpenCAD.  The hybrid framework consists of the following three main stages\footnote{For readers interested in the details of the hybrid framework, the pseudo-code is presented in Appendix \ref{app:code} (Alg. \ref{alg:llc}). The line numbers of the key steps are annotated in the description of Sect. \ref{sec:hybrid}.}. 

\noindent\textbf{Stage 1: $\boldsymbol{2d}$-LS.} Given a polynomial formula in which every relational operator is either `$<$' or `$>$', the first stage (Alg. \ref{alg:llc}, line \ref{line:2d_LS}--line \ref{line:assignment}) attempts to determine satisfiability using the 2D-LS algorithm. The choice to employ a local search algorithm first stems from its lightweight and incomplete nature. If it successfully finds a model, the solving process is usually quick. If unsuccessful, the subsequent stages will utilize complete solving algorithms. Furthermore, two outputs from 2D-LS are retained for later stages: the final cell-jump location and the number of cell-jump steps (Alg. \ref{alg:llc}, line \ref{line:assignment}). The final cell-jump location provides candidate variable assignments for the second stage, while the number of cell-jump steps helps estimate the number of unsatisfiable cells for the input formula, which will be used in the third stage.

\noindent\textbf{Stage 2: $\boldsymbol{2d}$-LS-Driven MCSAT.} The second stage (Alg. \ref{alg:llc}, line \ref{line:level_clause_set}--line \ref{line:MCSAT_end}) adopts an MCSAT framework as its foundational architecture. In the original MCSAT framework \cite{mcsat.Moura.2013}, assignments are made sequentially to each variable (typically set with default simple numerical values, e.g., 1, 0 or -1). In the absence of conflicts, this sequential assignment is computationally inexpensive. However, once a conflict occurs, an expensive explanation and backtracking procedure is triggered.

For satisfiable polynomial formulas, consider the scenario where the assignments to the first \(m\) variables are non-conflicting, yet a conflict arises when assigning the \((m+1)\)-th variable. In such a case, rather than performing the costly conflict-driven analysis, it is desirable to utilize the 2D-LS algorithm to efficiently find a model based on the partial assignment of the first \(m\) variables. To this end, we introduce a heuristic condition ``GoTo2DLS'' (Alg. \ref{alg:llc}, line \ref{line:heur_cond}) to evaluate the satisfiability potential of the input formula based on the current variable assignments. If this condition indicates a high probability of satisfiability, we invoke the 2D-LS algorithm on the formula after substituting the assignments of the first \(m\) variables.
\begin{enumerate}
    \item If the output is `SAT', we can combine current variable assignments in MCSAT with the model found by the 2D-LS algorithm to construct a model for the original input formula. 
    \item Otherwise, we use the final cell-jump location of the 2D-LS algorithm as candidate variable assignments for the unassigned variables in MCSAT. 
    (This choice is motivated by the understanding that the 2D-LS algorithm has executed several local modifications around this assignment without success, suggesting it is near critical constraints that could lead to conflicts. From a geometric perspective, there may be many unsatisfiable CAD cells near the cell where this assignment is located. Therefore, MCSAT can learn a lemma that covers multiple adjacent CAD cells from this assignment, ultimately enhancing the efficiency of the MCSAT process.)
\end{enumerate}
This stage employs a local search algorithm to efficiently solve the satisfiability of sub-formulas under partial assignments and guide variable assignments within the MCSAT framework, achieving a cohesive integration of the two methods. Consequently, we refer to this stage as ``2D-LS-driven MCSAT''. Furthermore, to prepare for the third stage, we collect learned clauses generated during the MCSAT procedure (Alg. \ref{alg:llc}, line \ref{line:learn_cl_1} and line \ref{line:learn_cl_2}), and update the estimation for the number of unsatisfiable cells (Alg. \ref{alg:llc}, line \ref{line:unsat_cell}) based on the number of cells visited by the 2D-LS algorithm. This estimation serves as a rough approximation, rather than providing the exact number of unsatisfiable cells.

\noindent\textbf{Stage 3: OpenCAD.} In the first two stages, we estimate the number of unsatisfiable cells for the input formula. The heuristic condition ``GoToOpenCAD'' is triggered when the estimation exceeds a predetermined threshold (Alg. \ref{alg:llc}, line \ref{line:openCAD_if}). If this condition is met, the algorithm transitions to this stage by invoking the OpenCAD procedure. OpenCAD is a complete method for deciding satisfiability of quantifier-free formulas that consists solely of strict inequality constraints. Integrating OpenCAD into our hybrid framework is motivated by the complementary strengths of the two complete methods: 
MCSAT operates as a CDCL-style search framework, showing superior performance in determining the satisfiability of formulas that are either satisfiable or unsatisfiable due to Boolean conflicts. In contrast, OpenCAD specializes in efficiently handling unsatisfiable formulas dominated by algebraic conflicts (refer to Example \ref{ex:OpenCAD} for a more detailed explanation).



\begin{example}\label{ex:OpenCAD}
The polynomial formula $f < 0$ with
\begin{align*}
    f = (x_1^2+x_2^2+x_3^2+x_4^2+x_5^2)^2 - 4(x_1^2 x_2^2+x_2^2 x_3^2+x_3^2 x_4^2+x_4^2 x_5^2+x_5^2 x_1^2)
\end{align*}
demonstrates two advantages of OpenCAD. First, OpenCAD outperforms standard CAD by exploiting strict inequalities: a full CAD of $\mathbb{R}^5$ produces approximately $2000$ cells, while OpenCAD generates fewer than $100$ by omitting lower-dimensional cells that are irrelevant for satisfiability. Second, this example favors algebraic approaches over MCSAT: MCSAT specializes in Boolean conflict learning but requires sampling points within each cell for algebraic constraints, making it less efficient than OpenCAD's direct geometric decomposition.
\end{example}

\begin{theorem}[Soundness and Completeness]
\label{thm:hybrid-sound-complete}
    The proposed hybrid framework is sound and complete.
\end{theorem}
The proof is given in Appendix~\ref{app:hybrid-proof}.

\section{Experiments}
\label{sec:exp}

We implement the hybrid framework with Mathematica 14 as a hybrid solver named HELMS\footnote{\url{https://github.com/ding-ty/HELMS-an-SMT-NRA-solver}}. We conduct experiments to evaluate the effectiveness and efficiency of HELMS on three classes of benchmarks, and compare its performance with baseline solvers and SOTA SMT solvers. Moreover, \reviewflag{stage variants are compared} to analyze the contribution of each stage in HELMS. \reviewflag{Per-instance results and supporting evidence for HELMS, 2D-LS and V1--V5 are available in the anonymous artifact repository, with code and configuration fingerprints.}

\subsection{Experiment Preliminaries}

In this paper, all experiments are conducted on 8-Core 11th Gen Intel(R) Core(TM) i7-11700@2.50GHz with 16GB of memory. Each solver is given a single attempt to solve each instance with a time limit of 60 seconds.

\noindent\textbf{Benchmarks:} We prepare the following three classes of benchmarks.
\begin{itemize}
    \item \textbf{SMTLIB.}We retain all $2050$ instances containing only strict polynomial inequalities from the SMT-LIB QF\_NRA collection\footnote{\url{https://smt-lib.org/}} ($20$ from \texttt{hong} and $2030$ from \texttt{meti-tarski}). After excluding these instance paths from our collected subset, we select one additional instance from each of $89$ eligible source directories, taking the path with the smallest SHA-256 hash in each directory. Selection is independent of solver answers and runtimes. We use their supplied strict-inequality transformations, which remove equality constraints and make weak inequalities strict, obtaining $2139$ instances in total. Since these transformations may change satisfiability, the additional instances are classified using results for the transformed formulas rather than inherited source labels. Appendix~\ref{app:smtlib-selection} lists the additional instances and their source-directory hierarchy.
    \item \textbf{RandInst.} Instances generated by function $\mathbf{rf}$ and parameters in \cite[Sect.~7.2]{ls2023}. We evaluate $100$ SAT instances. Unlike SMTLIB's abundance of unit clauses and linear constraints, the random instances are complicated in the Boolean structure (clauses contain two to five atoms) and highly nonlinear (polynomial total degrees range from $20$ to $30$).
    \item \textbf{SpecInst.} Instances from \cite[Sect.~5]{limbs2020}. This benchmark contains $21$ instances: $7$ SAT and $14$ UNSAT. These instances have particular mathematical properties that pose challenges for SOTA solvers. 
\end{itemize}

\noindent\textbf{Implementation:} The HELMS solver uses an MCSAT implementation based on LiMbS\footnote{\url{https://github.com/lihaokun/LiMbS}} and employs OpenCAD through Mathematica 14's built-in command \texttt{Generic\allowbreak Cylindrical\allowbreak Decomposition}. 

For an input polynomial formula, let $clausenum$ be the number of clauses, $polynum$ the number of polynomials, $n$ the number of variables, and $maxdeg$/ $mindeg$ the maximum/minimum degree of polynomials. Let $level$ and $maxlevel$ represent the current and historically maximum number of variables assigned by MCSAT, respectively.
\begin{itemize}
    \item \textbf{Parameters.} The HELMS solver is configured by the following five parameters. The maximum numbers of restarts for 2D-LS and 2D-LS-driven MCSAT are set to $MaxRestart_1=1$ and $MaxRestart_2=3$, respectively. The maximum number of cell-jumps is set to $MaxJump=10^5\cdot polynum\cdot n$. In addition, the number of randomly generated directions for cell-jumps is $2m=6$. Finally, the threshold for the ``GoToOpenCAD'' condition is set to $max\_numFailCells=0.1\cdot\min(polynum,maxdeg)\cdot n$.
    \item \textbf{Stage Switching.} The heuristic condition ``GoTo2DLS'' is $\min(0.4\cdot n,0.9\cdot maxlevel)<level<n-2$, and ``GoToOpenCAD'' is $numFailCells>max\_numFailCells$. Stage~1 is assigned a time limit of $2\cdot3^{mindeg/5-2}+2^{polynum/10-3/2}+2^{n/10-3/2}+clausenum/50-0.2$ seconds. Conflict-triggered switching from Stage~2 to Stage~3 additionally requires more than $20$ seconds in Stage~2 and $maxdeg>2$. 
\end{itemize}

\noindent\textbf{Competitors:} We prepare two classes of competitors for comparison.
\begin{itemize}
    \item \textbf{Baseline Solvers.} The first baseline solver is LS~\cite{ls2023}, a local search solver employing the original cell-jump operation. The second baseline solver is LiMbS~\cite{limbs2020}.
    \item \textbf{SOTA Solvers.} Four SOTA solvers from recent SMT Competitions (SMT-COMP) include Z3 (version 4.13.4), CVC5 (version 1.2.0), MathSAT5 (version 5.6.11) and Yices2 (version 2.6.5).
\end{itemize}

\noindent\textbf{Indicators:} Benchmark sizes are shown in parentheses. ``\#SAT'' (resp. ``\#UNSAT'') represents the number of instances correctly determined to be satisfiable (resp. unsatisfiable) by the solver, and ``\#ALL'' represents the sum of ``\#SAT'' and ``\#UNSAT''.

To guide the experimental analysis, we formulate research questions in Table~\ref{tab:rqs}. Evidence for the questions is presented as observations in Sects.~\ref{subsec:exp-performance}--\ref{subsec:stage}.
\begin{table}[t]
\vspace{-8mm}
\centering
\renewcommand{\arraystretch}{1.5}
\setlength{\tabcolsep}{2pt}
\caption{Research questions and corresponding empirical evidence.}\label{tab:rqs}
\begingroup
\papertablefont
\begin{tabularx}{\textwidth}{p{0.06\linewidth}|p{0.7\linewidth}|p{0.2\linewidth}}
\toprule
\textbf{ID} & \textbf{Research Question} & \textbf{Evidence}\\
\midrule
RQ1 & How does the performance of HELMS compare to SOTA SMT solvers in the context of SMT-NRA? & Observation 1, 6\\
RQ2 & Does the integration of MCSAT, local search, and OpenCAD successfully combine their individual advantages? & Observation 2, 3\\
RQ3 & What are the specific advantages of extending the local search range from 1D to 2D (2D-LS)? & Observation 4, 5\\
RQ4 & How do the three stages of HELMS contribute to performance, and are alternative stage combinations effective? & Observation 7, 8\\
\bottomrule
\end{tabularx}
\endgroup
\end{table}

\subsection{Number of Solved Instances}
\label{subsec:exp-performance}

Table~\ref{tab:3bench} compares the number of solved instances among baseline solvers, HELMS, and SOTA solvers, and Table~\ref{tab:components} presents those for HELMS and its components. Table~\ref{tab:2dls} compares the number of solved instances between 2D-LS and LS.

\begin{table}[t]
\vspace{-5mm}
\caption{The number of instances solved by HELMS, baseline solvers and SOTA solvers across three benchmarks.}\label{tab:3bench}
\centering
\begingroup
\papertablefont
\begin{tabular}{l c r r r r r r r}
\toprule
 & & \multicolumn{2}{@{}c@{}}{Baseline Solvers} & & \multicolumn{4}{@{}c@{}}{SOTA Solvers}\\
\cmidrule{3-4}\cmidrule{6-9}
\multicolumn{2}{@{}c@{}}{Benchmark} & {LS} & {LiMbS} & {\textbf{HELMS}} & {Z3} & {CVC5} & {MathSAT5} & {Yices2}\\
\midrule
\multirow{3}{*}{\benchsize{SMTLIB}{2139}} & {\#SAT} & \planchange{1513} & \planchange{\baselineupdate{1524}} & \planchange{1547} & \sotaupdate{\textbf{1552}} & \planchange{1520} & \sotaupdate{1308} & \sotaupdate{1544} \\
 & {\#UNSAT} & - & \planchange{574} & \planchange{\textbf{578}} & \planchange{576} & \planchange{575} & \sotaupdate{462} & \planchange{576} \\
 & {\#ALL} & \planchange{1513} & \planchange{\baselineupdate{2098}} & \planchange{2125} & \sotaupdate{\textbf{2128}} & \planchange{2095} & \sotaupdate{1770} & \sotaupdate{2120} \\ \hline
\benchsize{RandInst}{100} & {\#SAT} & \textbf{100} & \planchange{\baselineupdate{57}} & \textbf{100} & \sotaupdate{17} & 0 & \sotaupdate{0} & 23\\ \hline
\multirow{3}{*}{\benchsize{SpecInst}{21}} & {\#SAT} & 4 & \textbf{7} & \textbf{7} & \textbf{7} & 4 & 1 & \sotaupdate{3} \\
 & {\#UNSAT} & - & \planchange{\textbf{12}} & \planchange{10} & \sotaupdate{4} & 4 & 2 & 3\\
 & {\#ALL} & \planchange{4} & \planchange{\textbf{19}} & \planchange{17} & \sotaupdate{11} & 8 & 3 & \sotaupdate{6} \\ \hline
\multirow{3}{*}{\benchsize{Total}{2260}} & {\#SAT} & \planchange{1617} & \planchange{\baselineupdate{1588}} & \planchange{\textbf{1654}} & \sotaupdate{1576} & \planchange{1524} & \sotaupdate{1309} & \sotaupdate{1570} \\
 & {\#UNSAT} & - & \planchange{586} & \planchange{\textbf{588}} & \sotaupdate{580} & \planchange{579} & \sotaupdate{464} & \planchange{579} \\
 & {\#ALL} & \planchange{1617} & \planchange{\baselineupdate{2174}} & \planchange{\textbf{2242}} & \sotaupdate{2156} & \planchange{2103} & \sotaupdate{1773} & \sotaupdate{2149} \\ \hline
\bottomrule
\end{tabular}
\endgroup
\end{table}

\noindent\textbf{Observation 1. HELMS is competitive with SOTA solvers in terms of the number of solved instances.}
In the SMTLIB benchmark, as shown in Table~\ref{tab:3bench}, HELMS solves \planchange{$2125$} out of \planchange{$2139$} instances, \planchange{more than CVC5, MathSAT5 and Yices2, but fewer than Z3}. 
In the RandInst benchmark, HELMS solves all $100$ instances, whereas the best-performing SOTA solver, Yices2, solves only $23$, and neither CVC5 nor MathSAT5 solves any. This superiority
\planchange{derives from} the sign-flipping
capability of cell-jump operation for high-degree polynomials.
In the SpecInst benchmark, \reviewflag{HELMS solves \planchange{$17$} of the $21$ instances}. This performance benefits in part from its LiMbS-based MCSAT component.

\noindent\textbf{Observation 2. HELMS solves more instances than baseline solvers by integrating their strengths.}
In the SMTLIB benchmark, HELMS solves \planchange{$34$} more SAT instances than LS, with \planchange{$33$} solved by 2D-LS in Stage~1 and the remaining \planchange{one} by MCSAT in Stage~2. Additionally, \reviewflag{HELMS solves \planchange{$578$} UNSAT instances, compared with \planchange{$574$} for LiMbS}.
In the RandInst benchmark, LS and HELMS solve all $100$ instances, whereas LiMbS solves only \planchange{\baselineupdate{$57$}}. Conversely, in the SpecInst benchmark, \reviewflag{LiMbS and HELMS solve $19$ and \planchange{$17$} of the $21$ instances, respectively}, whereas LS solves only four. These results demonstrate that HELMS effectively combines the complementary strengths of LS and LiMbS.

\begin{table}[!t]
\centering
\captionsetup{font=small,justification=raggedright,singlelinecheck=false}
\caption{The number of instances solved by HELMS and its components (2D-LS, MCSAT and OpenCAD) across three benchmarks.}\label{tab:components}
\begingroup

\setlength{\tabcolsep}{1.2pt}
\begingroup
\papertablefont
\begin{tabular}{c c r r r r}
\toprule
 & & \multicolumn{3}{@{}c@{}}{HELMS Components} & \\
\cmidrule{3-5}
\multicolumn{2}{@{}c@{}}{Benchmark} & {2D-LS} & {LiMbS} & {OpenCAD} & {\textbf{HELMS}}\\
\midrule
\multirow{3}{*}{\benchsize{SMTLIB}{2139}} & {\#SAT} & \planchange{1546} & \planchange{\baselineupdate{1524}} & \planchange{\baselineupdate{1544}} & \planchange{\textbf{1547}} \\
 & {\#UNSAT} & - & \planchange{574} & \planchange{\baselineupdate{577}} & \planchange{\textbf{578}} \\
 & {\#ALL} & \planchange{1546} & \planchange{\baselineupdate{2098}} & \planchange{\baselineupdate{2121}} & \planchange{\textbf{2125}} \\ \hline
\benchsize{RandInst}{100} & {\#SAT} & \textbf{100} & \planchange{\baselineupdate{57}} & 0 & \textbf{100}\\ \hline
\multirow{3}{*}{\benchsize{SpecInst}{21}} & {\#SAT} & \textbf{7} & \textbf{7} & \planchange{2} & \textbf{7}\\
 & {\#UNSAT} & - & \planchange{\textbf{12}} & 8 & \planchange{10} \\
 & {\#ALL} & 7 & \planchange{\textbf{19}} & \planchange{10} & \planchange{17} \\ \hline
\multirow{3}{*}{\benchsize{Total}{2260}} & {\#SAT} & \planchange{1653} & \planchange{\baselineupdate{1588}} & \planchange{\baselineupdate{1546}} & \planchange{\textbf{1654}} \\
 & {\#UNSAT} & - & \planchange{586} & \planchange{\baselineupdate{585}} & \planchange{\textbf{588}} \\
 & {\#ALL} & \planchange{1653} & \planchange{\baselineupdate{2174}} & \planchange{\baselineupdate{2131}} & \planchange{\textbf{2242}} \\ \hline
\bottomrule
\end{tabular}
\endgroup
\endgroup
\end{table}

\noindent\textbf{Observation 3. HELMS effectively combines the advantages of its components.}
HELMS leverages 2D-LS for rapidly finding satisfying assignments and LiMbS for analyzing Boolean conflicts.
\reviewflag{Across the three benchmarks, HELMS solves \planchange{$1654$} SAT instances, maintaining the strong model-finding performance of 2D-LS at \planchange{$1653$}, while also solving \planchange{$588$} UNSAT instances, compared with \planchange{$586$} for LiMbS and \planchange{\baselineupdate{$585$}} for OpenCAD.
Consequently, HELMS solves \planchange{$2242$} instances in total, exceeding the strongest standalone component, LiMbS, by \planchange{\baselineupdate{$68$}}.
These results support the complementary roles of local model search, conflict-guided reasoning, and algebraic decomposition within HELMS.}

\begin{figure}[t]
\centering
\begin{minipage}[H]{0.4\textwidth}
\centering
\captionsetup{font=small,justification=raggedright,singlelinecheck=false}
\captionof{table}{The number of instances solved by LS and 2D-LS.}\label{tab:2dls}
\begingroup

\setlength{\tabcolsep}{2pt}
\renewcommand{\arraystretch}{1}
\noindent\makebox[\linewidth][l]{%
\begingroup
\papertablefont
\begin{tabular}{l r r}
\toprule
Benchmark & LS & 2D-LS \\
\midrule
\noalign{\vskip 2pt}
SMTLIB~(2139) & \planchange{1513} & \planchange{\textbf{1546}}\\[3pt]
\hline\noalign{\vskip 2pt}
RandInst~(100) & \textbf{100} & \textbf{100}\\[3pt]
\hline\noalign{\vskip 2pt}
SpecInst~(21) & 4 & \textbf{7}\\[3pt]
\hline\noalign{\vskip 2pt}
Total~(2260) & \planchange{1617} & \planchange{\textbf{1653}}\\[3pt]
\hline\noalign{\vskip 2pt}
\bottomrule
\end{tabular}
\endgroup%
}
\endgroup
\end{minipage}\hfill%
\begin{minipage}[H]{0.4\textwidth}
\centering
\includegraphics[width=0.8\linewidth]{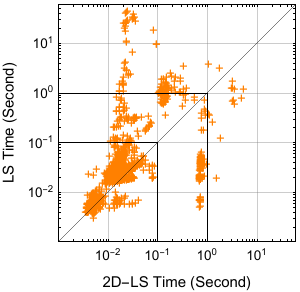}
\captionof{figure}{\small\planchange{Recorded solve times of 2D-LS and LS.}}\label{fig:rand-els-ls}\label{fig:experiment-runtime}
\end{minipage}
\vspace{-3mm}
\end{figure}

\noindent\textbf{Observation 4. \reviewflag{2D-LS solves more instances than LS.}} 
As shown in Table~\ref{tab:2dls}, \reviewflag{2D-LS solves \planchange{$1653$} instances in total, compared with \planchange{$1617$} for LS}. These results suggest that extending local search from one to two dimensions helps mitigate stagnation in local optima, enabling more satisfying assignments to be found.

\subsection{Efficiency}
\label{subsec:efficiency}

Fig.~\ref{fig:rand-els-ls} compares the efficiency between 2D-LS and LS. The recorded runtimes of HELMS and SOTA solvers are compared in the scatterplots in Appendix~\ref{app:helms-sota-scatter} (Fig.~\ref{fig:app-helms-sota-scatter}).

\noindent\textbf{Observation 5. \reviewflag{2D-LS exhibits greater efficiency compared to LS.}}
\reviewflag{Fig.~\ref{fig:rand-els-ls} plots recorded times for \planchange{$1616$} SAT instances solved by both methods within $60$ seconds. When both finish within \planchange{$1$} second, either can be faster, possibly reflecting randomized search and execution overheads. Among \planchange{$107$} cases with at least one recorded time exceeding \planchange{$1$} second, 2D-LS is faster on \planchange{$93$} and LS on \planchange{$14$}. This pattern is consistent with planar search alleviating prolonged stagnation.}

\noindent\textbf{Observation 6. HELMS is competitive to SOTA solvers in terms of runtime.}
On jointly solved instances, HELMS and SOTA solvers have median recorded runtimes of the same order of magnitude, while HELMS is generally slower. HELMS is a Mathematica prototype, whereas the competitors are mature, extensively optimized SMT solvers; these implementation differences may contribute to the runtime gap. Fig.~\ref{fig:app-helms-sota-scatter} also shows that HELMS is relatively more competitive on SAT than on UNSAT instances. This pattern is consistent with efficient model search by 2D-LS in Stage~1, which can find satisfying assignments before later stages are needed.

\subsection{Evaluation of HELMS Stages}
\label{subsec:stage}

To analyze the contribution of each stage in HELMS, Table~\ref{tab:phase} presents the distribution of terminal stages and compares HELMS with its variants.

\begin{table}[t]
\caption{\planchange{The number of instances solved at each HELMS stage and by its variants.}}
\label{tab:phase}\label{tab:v1-v5}
\centering
\par\smallskip
\begingroup
\setlength{\tabcolsep}{2pt}
\renewcommand{\arraystretch}{1.05}

\begingroup
\papertablefont
\begin{tabular}{@{}l c r r r r r r r r r@{}}
\toprule
 & & \multicolumn{3}{c}{Stages of HELMS}
 & & \multicolumn{5}{c}{HELMS Variants}\\
\cmidrule(lr){3-5}\cmidrule(lr){7-11}
\multicolumn{2}{@{}c@{}}{Benchmark}
 & {Stage 1}
 & {Stage 2}
 & {Stage 3}
 & HELMS
 & {V1}
 & {V2}
 & {V3}
 & {V4}
 & {V5}\\
\midrule

\multirow{3}{*}{\benchsize{SMTLIB}{2139}}
 & {\#SAT}
 & \planchange{1546}
 & \planchange{1}
 & \planchange{0}
 & \planchange{1547}
 & \planchange{1546}
 & \planchange{1546}
 & \planchange{1546}
 & \planchange{1547}
 & \planchange{1546}\\
 & {\#UNSAT}
 & \planchange{0}
 & \planchange{573}
 & \planchange{5}
 & \planchange{578}
 & \planchange{575}
 & \planchange{577}
 & \planchange{574}
 & \planchange{576}
 & \planchange{575}\\
 & {\#ALL}
 & \planchange{1546}
 & \planchange{574}
 & \planchange{5}
 & \planchange{2125}
 & \planchange{2121}
 & \planchange{2123}
 & \planchange{2120}
 & \planchange{2123}
 & \planchange{2121}\\
\midrule

\benchsize{RandInst}{100}
 & {\#SAT}
 & 100
 & 0
 & 0
 & 100
 & 100
 & 100
 & 100
 & 15
 & 100\\
\midrule

\multirow{3}{*}{\benchsize{SpecInst}{21}}
 & {\#SAT}
 & \planchange{7}
 & \planchange{0}
 & 0
 & 7
 & \planchange{5}
 & \planchange{4}
 & 7
 & \planchange{5}
 & 7\\
 & {\#UNSAT}
 & 0
 & \planchange{7}
 & \planchange{3}
 & \planchange{10}
 & \planchange{11}
 & 8
 & \planchange{9}
 & \planchange{13}
 & \planchange{11}\\
 & {\#ALL}
 & \planchange{7}
 & \planchange{7}
 & \planchange{3}
 & \planchange{17}
 & \planchange{16}
 & \planchange{12}
 & \planchange{16}
 & \planchange{18}
 & \planchange{18}\\
\midrule

\multirow{3}{*}{\benchsize{Total}{2260}}
 & {\#SAT}
 & \planchange{1653}
 & \planchange{1}
 & \planchange{0}
 & \planchange{1654}
 & \planchange{1651}
 & \planchange{1650}
 & \planchange{1653}
 & \planchange{1567}
 & \planchange{1653}\\
 & {\#UNSAT}
 & \planchange{0}
 & \planchange{580}
 & \planchange{8}
 & \planchange{588}
 & \planchange{586}
 & \planchange{585}
 & \planchange{583}
 & \planchange{589}
 & \planchange{586}\\
 & {\#ALL}
 & \planchange{1653}
 & \planchange{581}
 & \planchange{8}
 & \planchange{2242}
 & \planchange{2237}
 & \planchange{2235}
 & \planchange{2236}
 & \planchange{2156}
 & \planchange{2239}\\
\hline
\bottomrule
\end{tabular}
\endgroup%
\endgroup
\vspace{-5mm}
\end{table}

\noindent\textbf{Observation 7. Each stage of HELMS makes a contribution to the overall performance.}
Stage~1 (2D-LS) solves \planchange{$1653$} SAT instances. It uses local search to seek models quickly, especially for SAT problems with highly nonlinear constraints; all $100$ RandInst instances terminate in this stage.
Stage~2 (2D-LS-driven MCSAT) solves \planchange{$1$} SAT and \planchange{$580$} UNSAT instances. Its conflict-guided reasoning complements model search on the remaining instances, particularly UNSAT cases dominated by Boolean conflicts and instances with special mathematical properties, such as those in SpecInst.
Stage~3 (OpenCAD) solves \planchange{$8$} UNSAT instances: \planchange{$5$} in SMTLIB and \planchange{$3$} in SpecInst (Table~\ref{tab:phase}). This stage handles residual UNSAT instances with complex algebraic conflicts, for which direct algebraic decomposition complements the search performed in the preceding stages. 

We conduct an ablation study to evaluate HELMS under different combinations of its stages. The following five variants are considered:
\begin{itemize}
    \item V1: Remove Stage~1 from HELMS.
    \item V2: Remove Stage~2 from HELMS, retaining 2D-LS followed by OpenCAD.
    \item V3: Remove Stage~3 from HELMS.
    \item V4: Remove Stage~1 and disable 2D-LS in Stage~2.
    \item V5: Disable the use of final cell-jump locations from 2D-LS as candidate variable assignments for MCSAT.
\end{itemize}
We retain the relative order of the remaining stages in each variant, following the design rationale of HELMS. Local search is placed first to quickly find SAT models before invoking more systematic reasoning. Although OpenCAD is complete, direct algebraic decomposition can be computationally expensive. MCSAT is therefore prioritized, with OpenCAD reserved for difficult algebraic instances that remain unresolved. This motivates the stage order Stage~1 $\rightarrow$ Stage~2 $\rightarrow$ Stage~3.

\noindent\textbf{Observation 8. \reviewflag{HELMS outperforms other stage combinations.}}
\reviewflag{HELMS solves \planchange{$2242$} instances overall, and its additional successes illustrate the roles of its components (see Table~\ref{tab:v1-v5}). There are \planchange{$3$} SAT instances missed by V1 that are solved in Stage~1, supporting the use of an initial local-search phase. Likewise, the \planchange{$85$} RandInst instances missed by V4 all terminate in Stage~1 of HELMS, showing how initial model search handles high-degree SAT instances before entering Stage~2. The \planchange{$3$} UNSAT instances missed by V2 are resolved in Stage~2, illustrating how MCSAT complements local search and OpenCAD. In addition, the  \planchange{$7$} UNSAT instances missed by V3 are resolved by OpenCAD, supporting its role as an algebraic fallback. Together, these cases illustrate the complementary roles of early model search, conflict-guided reasoning, and algebraic decomposition in the recorded configurations.}

\section{Conclusion}
\label{sec:conclu}
In this paper, we present HELMS, a hybrid SMT-NRA solver integrating 2D-LS, MCSAT, and OpenCAD for quantifier-free formulas with strict inequalities.
First, HELMS invokes 2D-LS once in an attempt to quickly find a satisfying assignment. Then, HELMS adopts an MCSAT framework as a foundational architecture, with appropriately triggering 2D-LS to suggest variable assignments for MCSAT. Finally, HELMS utilizes OpenCAD to handle unsatisfiable formulas that are dominated by algebraic conflicts.
Experimental results demonstrate its effectiveness on various benchmarks, particularly for polynomial formulas with high-degree constraints or complex Boolean structures, or particular mathematical properties.

{\small\subsubsection*{\ackname}
The authors are grateful to the reviewers for their valuable comments and constructive suggestions. 
This work was supported by the National Key R\&D Program of China under grant Nos.~2022YFA1005102 and 2024YFA1014000, and by the National Natural Science Foundation of China under grant No.~62602614.}

\bibliographystyle{splncs04}  
\bibliography{paper}  

@article{chen2025geometry,
  title={The Geometry of Cylindrical Algebraic Decomposition},
  author={Chen, Rizeng},
  journal={ACM Communications in Computer Algebra},
  volume={58},
  number={3},
  pages={67--71},
  year={2025},
  publisher={ACM New York, NY, USA}
}

@article{XIA2002633,
title = {A complete algorithm for counting real solutions of polynomial systems of equations and inequalities},
journal = {Computers \& Mathematics with Applications},
volume = {44},
number = {5},
pages = {633-642},
year = {2002},
issn = {0898-1221},
doi = {https://doi.org/10.1016/S0898-1221(02)00178-5},
url = {https://www.sciencedirect.com/science/article/pii/S0898122102001785},
author = {Bican Xia and Xiaorong Hou}
}

@InProceedings{10.1007/978-3-031-99984-0_6,
author="Lipparini, Enrico
and Hader, Thomas
and Irfan, Ahmed
and Graham-Lengrand, St{\'e}phane",
editor="Barrett, Clark
and Waldmann, Uwe",
title="Boosting MCSat Modulo Nonlinear Integer Arithmetic via Local Search",
booktitle="Automated Deduction -- CADE 30",
year="2025",
publisher="Springer Nature Switzerland",
address="Cham",
pages="95--115",
isbn="978-3-031-99984-0"
}

@inproceedings{akinyele2013using,
  title={{Using SMT Solvers to Automate Design Tasks for Encryption and Signature Schemes}},
  author={Akinyele, Joseph A and Green, Matthew and Hohenberger, Susan},
  booktitle={Proceedings of the 2013 ACM SIGSAC conference on Computer \& communications security},
  pages={399--410},
  year={2013}
}

@inproceedings{gundaram2021performance,
  title={{Performance of Various SMT Solvers in Cryptanalysis}},
  author={Gundaram, Praveen Kumar and Tentu, Appala Naidu and Muppalaneni, Naresh Babu},
  booktitle={2021 international conference on computing, communication, and intelligent systems (ICCCIS)},
  pages={298--303},
  year={2021},
  organization={IEEE}
}

@inproceedings{baelde2025smt,
  title={{SMT-Based Automation for Overwhelming Truth}},
  author={Baelde, David and Delaune, St{\'e}phanie and Riou, Stanislas},
  booktitle={38th IEEE Computer Security Foundations Symposium},
  year={2025},
  organization={IEEE}
}

@inproceedings{leino2013automating,
  title={{Automating Theorem Proving with SMT}},
  author={Leino, K Rustan M},
  booktitle={International Conference on Interactive Theorem Proving},
  pages={2--16},
  year={2013},
  organization={Springer}
}

@incollection{keller2019smtcoq,
  title={{SMTCoq: Mixing Automatic and Interactive Proof Technologies}},
  author={Keller, Chantal},
  booktitle={Proof Technology in Mathematics Research and Teaching},
  pages={73--90},
  year={2019},
  publisher={Springer}
}

@article{weber2011smt,
  title={{SMT Solvers: New Oracles for the HOL Theorem Prover}},
  author={Weber, Tjark},
  journal={International Journal on Software Tools for Technology Transfer},
  volume={13},
  number={5},
  pages={419--429},
  year={2011},
  publisher={Springer}
}

@InProceedings{mcsat.Moura.2013,
author="de Moura, Leonardo
and Jovanovi{\'{c}}, Dejan",
editor="Giacobazzi, Roberto
and Berdine, Josh
and Mastroeni, Isabella",
title="A Model-Constructing Satisfiability Calculus",
booktitle="Verification, Model Checking, and Abstract Interpretation",
year="2013",
publisher="Springer Berlin Heidelberg",
address="Berlin, Heidelberg",
pages="1--12",
isbn="978-3-642-35873-9"
}

@article{limbs2020,
  author       = {Haokun Li and
                  Bican Xia},
  title        = {{Solving Satisfiability of Polynomial Formulas by Sample-Cell Projection}},
  journal      = {CoRR},
  volume       = {abs/2003.00409},
  year         = {2020},
  url          = {https://arxiv.org/abs/2003.00409},
  eprinttype    = {arXiv},
  eprint       = {2003.00409},
  bibsource    = {dblp computer science bibliography, https://dblp.org}
}

@INPROCEEDINGS{deep2021,
  author={Zhang, Xindi and Li, Bohan and Cai, Shaowei},
  booktitle={2024 IEEE/ACM 46th International Conference on Software Engineering (ICSE)}, 
  title={{Deep Combination of CDCL(T) and Local Search for Satisfiability Modulo Non-Linear Integer Arithmetic Theory}}, 
  year={2024},
  volume={},
  number={},
  pages={1534-1546},
  doi={10.1145/3597503.3639105}}

@InProceedings{ls2023,
author="Li, Haokun
and Xia, Bican
and Zhao, Tianqi",
editor="Enea, Constantin
and Lal, Akash",
title={{Local Search for Solving Satisfiability of Polynomial Formulas}},
booktitle="Computer Aided Verification",
year="2023",
publisher="Springer Nature Switzerland",
address="Cham",
pages="87--109",
isbn="978-3-031-37703-7"
}

@InProceedings{realcaiLS,
author="Wang, Zhonghan
and Zhan, Bohua
and Li, Bohan
and Cai, Shaowei",
editor="Dimitrova, Rayna
and Lahav, Ori
and Wolff, Sebastian",
title={{Efficient Local Search for Nonlinear Real Arithmetic}},
booktitle="Verification, Model Checking, and Abstract Interpretation",
year="2024",
publisher="Springer Nature Switzerland",
address="Cham",
pages="326--349",
isbn="978-3-031-50524-9"
}

@INPROCEEDINGS{realmulcaiLS,
  author={Li, Bohan and Cai, Shaowei},
  booktitle={2023 Formal Methods in Computer-Aided Design (FMCAD)}, 
  title={{Local Search for SMT on Linear and Multi-linear Real Arithmetic}}, 
  year={2023},
  volume={},
  number={},
  pages={1-10},
  doi={10.34727/2023/isbn.978-3-85448-060-0_25}}

@inproceedings{hybSAT2008,
  title={{Improvements to Hybrid Incremental SAT Algorithms}},
  author={Florian Letombe and Joao Marques-Silva},
  booktitle={International Conference on Theory and Applications of Satisfiability Testing},
  year={2008},
  url={https://api.semanticscholar.org/CorpusID:16633191}
}

@InProceedings{z32008,
author={de Moura, Leonardo
and Bj{\o}rner, Nikolaj},
editor={Ramakrishnan, C. R.
and Rehof, Jakob},
title={{Z3: An Efficient SMT Solver}},
booktitle={Tools and Algorithms for the Construction and Analysis of Systems},
year={2008},
publisher={Springer Berlin Heidelberg},
address={Berlin, Heidelberg},
pages={337--340},
isbn={978-3-540-78800-3}
}

@InProceedings{cvc52022,
author="Barbosa, Haniel
and Barrett, Clark
and Brain, Martin
and Kremer, Gereon
and Lachnitt, Hanna
and Mann, Makai
and Mohamed, Abdalrhman
and Mohamed, Mudathir
and Niemetz, Aina
and N{\"o}tzli, Andres
and Ozdemir, Alex
and Preiner, Mathias
and Reynolds, Andrew
and Sheng, Ying
and Tinelli, Cesare
and Zohar, Yoni",
editor="Fisman, Dana
and Rosu, Grigore",
title={{cvc5: A Versatile and Industrial-Strength SMT Solver}},
booktitle="Tools and Algorithms for the Construction and Analysis of Systems",
year="2022",
publisher="Springer International Publishing",
address="Cham",
pages="415--442",
isbn="978-3-030-99524-9"
}

@InProceedings{mathSAT5,
author="Cimatti, Alessandro
and Griggio, Alberto
and Schaafsma, Bastiaan Joost
and Sebastiani, Roberto",
editor="Piterman, Nir
and Smolka, Scott A.",
title={{The MathSAT5 SMT Solver}},
booktitle="Tools and Algorithms for the Construction and Analysis of Systems",
year="2013",
publisher="Springer Berlin Heidelberg",
address="Berlin, Heidelberg",
pages="93--107",
isbn="978-3-642-36742-7"
}

@InProceedings{Yices2,
author="Dutertre, Bruno",
editor="Biere, Armin
and Bloem, Roderick",
title={{Yices 2.2}},
booktitle="Computer Aided Verification",
year="2014",
publisher="Springer International Publishing",
address="Cham",
pages="737--744",
isbn="978-3-319-08867-9"
}

@inproceedings{smtlib,
   author = {Clark Barrett and Aaron Stump and Cesare Tinelli},
   title = {{The {SMT-LIB} Standard -- Version 2.0}},
   booktitle = {Proceedings of the  $8^{th}$ International Workshop on Satisfiability Modulo Theories (SMT '10)},
   month = jul,
   year = {2010},
   note = {Edinburgh, Scotland},
   url = {http://theory.stanford.edu/~barrett/pubs/BST10.pdf}
}

@article{Alur2011FormalVO,
  title={{Formal Verification of Hybrid Systems}},
  author={Rajeev Alur},
  journal={2011 Proceedings of the Ninth ACM International Conference on Embedded Software (EMSOFT)},
  year={2011},
  pages={273-278},
  url={https://api.semanticscholar.org/CorpusID:14278725}
}

@inproceedings{chen2018control,
  title={{Control Flow-Guided SMT Solving for Program Verification}},
  author={Chen, Jianhui and He, Fei},
  booktitle={Proceedings of the 33rd ACM/IEEE International Conference on Automated Software Engineering},
  pages={351--361},
  year={2018}
}

@article{cimatti2013smt,
  title={{SMT-Based Scenario Verification for Hybrid Systems}},
  author={Cimatti, Alessandro and Mover, Sergio and Tonetta, Stefano},
  journal={Formal Methods in System Design},
  volume={42},
  pages={46--66},
  year={2013},
  publisher={Springer}
}

@article{faure2024methodology,
  author={Faure-Gignoux, Anthony and Delmas, Kevin and Gauffriau, Adrien and Pagetti, Claire},
  journal={IEEE Embedded Systems Letters}, 
  title={{Methodology for Formal Verification of Hardware Safety Strategies Using SMT}}, 
  year={2024},
  volume={16},
  number={4},
  pages={381-384},
  doi={10.1109/LES.2024.3439859}}

@inproceedings{li2010scalable,
  title={{Scalable SMT-Based Verification of GPU Kernel Functions}},
  author={Li, Guodong and Gopalakrishnan, Ganesh},
  booktitle={Proceedings of the eighteenth ACM SIGSOFT international symposium on Foundations of software engineering},
  pages={187--196},
  year={2010}
}

@article{beyer2018unifying,
  title={{A Unifying View on SMT-Based Software Verification}},
  author={Beyer, Dirk and Dangl, Matthias and Wendler, Philipp},
  journal={Journal of automated reasoning},
  volume={60},
  number={3},
  pages={299--335},
  year={2018},
  publisher={Springer}
}

@article{hardVeri2016,
  title={{Applying SMT-Based Verification to Hardware/Software Partitioning in Embedded Systems}},
  author={Trindade, Alessandro B and Cordeiro, Lucas C},
  journal={Design Automation for Embedded Systems},
  volume={20},
  pages={1--19},
  year={2016},
  publisher={Springer}
}

@incollection{collins1975quantifier,
  title={{Quantifier Elimination for Real Closed Fields by Cylindrical Algebraic Decompostion}},
  author={Collins, George E},
  booktitle={Automata Theory and Formal Languages},
  pages={134--183},
  year={1975},
  publisher={Springer}
}

@incollection{tarski1998decision,
  title={{A Decision Method for Elementary Algebra and Geometry}},
  author={Tarski, Alfred},
  booktitle={Quantifier elimination and cylindrical algebraic decomposition},
  pages={24--84},
  year={1998},
  publisher={Springer}
}

@inproceedings{han2014constructing,
  title={{Constructing Fewer Open Cells by GCD Computation in CAD Projection}},
  author={Han, Jingjun and Dai, Liyun and Xia, Bican},
  booktitle={Proceedings of the 39th International Symposium on Symbolic and Algebraic Computation},
  pages={240--247},
  year={2014}
}

@String{Computing = "Computing" }

@String{Computer = "{IEEE} Computer" }

@String{Springer = "Springer-Verlag" }

\newpage
\appendix
\renewcommand{\thesection}{\Alph{section}}%

\section{Hybrid Framework Pseudo-Code}
\label{app:code}

Alg. \ref{alg:llc} is a pseudo-code for our hybrid framework, where the black lines correspond to the MCSAT procedure, and the blue lines represent the extensions introduced by the hybrid framework. 


\begin{algorithm}[htbp]
\fontsize{6}{7}\selectfont
\SetAlgoNlRelativeSize{-1}
\caption{The Hybrid Framework}
\label{alg:llc}
\DontPrintSemicolon
\LinesNumbered
\KwIn{$F$, a polynomial formula with variables $x_1,\ldots,x_n$ such that the relational operator of every atom is `$<$' or `$>$'; $MaxRestart_1$, $MaxRestart_2$, $MaxJump$, $m$ and $max\_numFailCells$, five positive integers
}
\KwOut{`SAT' or `UNSAT'}
\SetKwFunction{FDecide}{Decide}
\SetKwFunction{FCheckNoConflict}{CheckNoConflict}
\SetKwFunction{FVariables}{Variables}
\SetKwFunction{FSimplifiedCAD}{OpenCAD}
\SetKwFunction{FSampleCellProjection}{SampleCellProjection}
\SetKwFunction{FMCSATD}{MCSATDriven2DLS}
\SetKwFunction{FChoose}{Choose}
\SetKwFunction{FSolve}{Solve}
\SetKwFunction{FProj}{Proj}

{\color{blue} $(status,\alpha,numJump)\gets2D\text{-LS}(F,MaxRestart_1,MaxJump,m)$
}\label{line:2d_LS}

{\color{blue} \lIf{$status=${\rm`SAT'}}{\Return{\rm`SAT'}}}

{\color{blue} $assignment\gets\alpha$, $numFailCells\gets numJump$, $maxlevel\gets0$}\label{line:assignment}


\lFor{$i$ from $1$ to $n$}{
$CS_i\gets \{c\in\clauses(F)\mid\level(c)=i\}$\label{line:level_clause_set}
}

$level\gets1$, $M\gets []$, $learnClauses\gets\emptyset$\label{line:trail_1}

\While{$\true$}{
    \uIf{$\newvalue(CS_{level},M)=\true$\label{line:hyb-value-true}}
        {
        \If{$assignment[x_{level}]$ not in $\FSolve(M)$}{
            $assignment[x_{level}]\gets$ one element in $\FSolve(M)$
        }
        $M \gets [M;x_{level}\mapsto assignment[x_{level}]]$\label{line:trail_2}

        $level++$

        {\color{blue} $maxlevel\gets\max(level,maxlevel)$}

        \lIf{$level>n$}{\Return{\rm`SAT'}}\label{line:hyb-sat}
        {\color{blue}\If{GoTo2DLS\label{line:heur_cond}}
        {
            $(status,\alpha,numJump)\gets\allowbreak2D\text{-LS}(F\mid_{x_1=assignment[x_1],\ldots,x_{level-1}=assignment[x_{level-1}]},MaxRestart_2,MaxJump,m)$\label{line:mcsat-2dls}

            \lIf{$status=${\rm`SAT'}}{\Return{\rm`SAT'}}
            \Else{
                \lFor{i from $level$ to $n$}{$assignment[x_{i}]\gets\alpha[x_i]$}
                
                $numFailCells\gets numFailCells+numJump$ \label{line:hyb-2dLS-end}
            }
        }}
    }
    \Else{\label{line:hyb-else}
        \lIf{there exists a clause $c$ in $CS_{level}$ such that $\newvalue(c,M) =\false$}{
            $mcstatus\gets(\text{`UNSAT'},c)$
        }
        \lElseIf{there exists a clause $c$ in $CS_{level}$ such that $\newvalue(c,M)=\newundef$ and only one literal $\ell$ in clause $c$ such that $\newvalue(\ell,M)=\newundef$}{
            $mcstatus\gets (\text{`Propagate'},\ell,c)$
        }
        \Else{
            choose a clause $c$ in $CS_{level}$ such that $\newvalue(c,M) =\newundef$
            
            choose a literal $\ell$ in $c$ such that $\newvalue(\ell,M) =\newundef$
            
            $mcstatus\gets(\text{`Decide'},\ell,c)$ \label{line:hyb-decide}
        }

        \If{$mcstatus[1]\ne$ {\rm`UNSAT'} \label{line:hyb-neq unsat}}{
            $\ell\gets mcstatus$[2], $c\gets mcstatus$[3]

            \uIf{$\Cons(\ell,M)$ \label{line:hyb-consistent}}{
                \lIf{$mcstatus[1]=$ {\rm`Decide'}}{
                    $M \gets [M; \ell]$\label{line:trail_3}
                }   
                \lElseIf{$mcstatus[1]=$ {\rm `Propagate'}}{
                    $M \gets [M; c\to\ell]$\label{line:trail_4}
                }
            }
            \Else{
                $minCore\gets$ minimal conflicting core of  $M$ and $\ell$ on the level $level$
                
                $cell \gets \explain(minCore,(assignment[x_1],\ldots,assignment[x_{level-1}]))$
                
                $lemma \gets \neg \left(cell \land (\bigwedge_{\ell'\in minCore} \ell') \land \ell\right)$
                
                {\color{blue}$learnClauses \gets learnClauses \cup \{lemma\}$\label{line:learn_cl_1}}
                
                $CS_{level} \gets CS_{level}\cup\{lemma\}$
                
                $M \gets [M; lemma \to \neg \ell]$\label{line:trail_5}

                \If{$mcstatus[1]=$ {\rm`Propagate'}\label{line:hyb-status-propagate}}{
                    {\color{blue} $numFailCells\gets numFailCells+1$}\label{line:unsat_cell}
                    
                    $mcstatus\gets ($`UNSAT'$,c)$ \label{line:hyb-consistent-end}
                }
            }
        }

        \If{$mcstatus[1]=$ {\rm`UNSAT'} \label{line:hyb-unsat}}{
            $c\gets mcstatus[2]$, $lemma \gets\resolve(c,M,level)$ 
            
            \lIf{$lemma$ is empty}{\Return{\rm`UNSAT'}}
            
            {\color{blue}$learnClauses \gets learnClauses \cup \{lemma\}$\label{line:learn_cl_2}}
    
            \uIf{$x_{level}$ appears in $lemma$}
            {
                $CS_{level} \gets CS_{level} \cup \{lemma\}$

                $\ell^{*}\gets$ the last decided literal in $M$ satisfying $\ell^{*}\in\atoms(lemma)$
                
                delete the decided literal $\ell^{*}$ and all subsequent terms from $M$
                
            }
    
            \Else{
                $tmpLevel\gets$ maximal $i$ of $x_{i}$ appearing in $lemma$
                
                $CS_{tmpLevel} \gets CS_{tmpLevel} \cup \{lemma\}$
                
                delete the assignment $x_{tmpLevel}\mapsto assignment[x_{tmpLevel}]$ and all subsequent terms from $M$
                
                $level\gets tmpLevel$\label{line:MCSAT_end}

                }
                
        }

    }
    {
        \color{blue} \If{$numFailCells > max\_numFailCells$\label{line:openCAD_if}}
        {\Return{{\rm OpenCAD}{$(F)$}\label{line:openCAD}}}
    }
}
    \vspace{-1mm}
\end{algorithm}

\textbf{Introduction to the MCSAT Procedure.} To be reader-friendly, we provide an introduction to the original MCSAT framework, \ie the black lines in Alg. \ref{alg:llc}. Lines \ref{line:hyb-value-true}--\ref{line:hyb-sat} assign values to variables when all clauses at the current level evaluate to true. If all variables are successfully assigned, the algorithm can terminate with `SAT'. 
If the assignment fails, line \ref{line:hyb-else}--line \ref{line:hyb-decide}, line \ref{line:hyb-status-propagate}--line \ref{line:hyb-consistent-end} yield either `Decide', `Propagate', or `UNSAT' statuses. For `Decide' or `Propagate' statuses, black lines in line \ref{line:hyb-consistent}--line \ref{line:hyb-consistent-end} verify the consistency of the decided or propagated literal with the trail $M$.
If an inconsistency is detected, a lemma is generated that represents an unsatisfiable region, expressed in the form of a clause to be excluded from future searches.  This process involves using a single-cell projection operator \cite{limbs2020} to project an assignment onto the lemma. Additionally, if an inconsistency occurs during the 'Propagate' status, the algorithm updates and transitions to `UNSAT'.
For `UNSAT' statuses, black lines in line \ref{line:hyb-unsat}--line \ref{line:MCSAT_end} carry out lemma learning (an empty lemma makes the algorithm terminate with `UNSAT'), and then executes non-chronological backtracking driven by the lemma. This process repeats iteratively until termination.

\section{Proof of Theorem~\ref{thm:hybrid-sound-complete}}
\label{app:hybrid-proof}

\begin{proof}
For soundness, a model returned by 2D-LS satisfies $F$, after combining it with the fixed partial assignment in Stage~2 when necessary. Other answers follow from the soundness of MCSAT~\cite{mcsat.Moura.2013,limbs2020} or OpenCAD~\cite{han2014constructing}, which is applied to the original $F$.

For completeness, each 2D-LS call terminates under its finite budgets. Unsuccessful calls supply only candidate decision values, which MCSAT accepts only when consistent with its current trail. Such choices preserve the conflict explanation rules and their finite basis, as well as learning and backtracking. Hence the termination and completeness arguments for MCSAT~\cite[Theorem~1]{mcsat.Moura.2013} and its sample-cell instantiation~\cite[Theorem~4]{limbs2020} remain applicable. If Stage~3 is not reached, Stage~2 therefore eventually decides $F$.

If Stage~3 is reached, OpenCAD decides the original $F$. Since the model set of $F$ is open, any nonempty model set contains a full-dimensional sign-invariant CAD cell. Thus OpenCAD's completeness for sampling full-dimensional regions~\cite[Theorem~3]{han2014constructing} ensures that it finds a model whenever one exists and otherwise establishes unsatisfiability. In either case, the framework terminates with the correct answer.
\end{proof}

\section{SMTLIB Instance Selection}
\label{app:smtlib-selection}

The tree lists all $89$ additional source files alongside
the two retained original collections.
Nested paths are compressed using slashes, and the common
\texttt{.smt2} extension is omitted.

\begin{lstlisting}[
  basicstyle=\ttfamily\scriptsize,
  columns=fullflexible,
  keepspaces=true,
  breaklines=true,
  breakatwhitespace=false,
  showstringspaces=false
]
QF_NRA/
|-- 20161105-Sturm-MBO/
|   `-- mbo_E23E24
|-- 20161105-Sturm-MGC/
|   `-- mgc_06
|-- 20170501-Heizmann-UltimateInvariantSynthesis/
|   `-- overflow1.i_3_8_2.bpl_9
|-- 20180501-Economics-Mulligan/
|   `-- MulliganEconomicsModel0085a
|-- 20200911-Pine/
|   `-- 1599122164633784000
|-- 20211101-Geogebra/
|   `-- IsoRightTriangle-Bottema8_1b
|-- 20220314-Uncu/
|   |-- Polynomial_Proof_Calls/ProveIneq_JSC_SUMCracker_19b
|   |-- Rational_Function_Proof_Calls_With_Basic_Denominator_Simplifications/BDC_ProveIneq_ISSAC10_ex3f
|   `-- Rational_Function_Proof_Calls_With_Disjunctive_Denominator_Simplifications/DDC_ProveIneq_ISSAC10_ex1f
|-- 20240407-pPDA-Chiari-Pontiggia-Winkler/
|   |-- certificate_hints/mod5_POMC_hints
|   |-- certificate_no_hints/schelling_POMC_no_hints
|   `-- past/sqrt2by2_bern_PRAY
|-- LassoRanker/
|   |-- CooperatingT2/nested.t2.c_Iteration4_Loop_3-pieceTemplate
|   |-- SV-COMP/NoriSharma-2013FSE-Fig8_true-termination.c_Iteration1_Loop_7-phaseTemplate
|   `-- Ultimate/Collatz.bpl_Iteration1_Lasso_6-nestedTemplate
|-- UltimateAutomizer/
|   `-- InVarSynth_DivisibilityInterpolantRequired02-ForwardSuccess.bpl_Iteration1_0
|-- hong/ [20 retained original instances]
|-- hycomp/
|   `-- ball_count_2d_plain.01.seq_lazy_lemmas_global_10
|-- kissing/
|   `-- kissing_4_9
|-- meti-tarski/ [2030 original instances; additions below]
|   |-- Arthan/
|   |   |-- 1A/Arthan1A-chunk-0019
|   |   |-- 1C/Arthan1C-chunk-0005
|   |   |-- KM2/ArthanKM2-chunk-0008
|   |   `-- M2/ArthanM2-chunk-0007
|   |-- CMOS/
|   |   `-- CMOS-opamp-chunk-0118
|   |-- CONVOI2/
|   |   `-- CONVOI2-chunk-0016
|   |-- Chua/
|   |   |-- 1/IL/L/Chua-1-IL-L-chunk-0045
|   |   |-- 1/IL/U/Chua-1-IL-U-chunk-0072
|   |   |-- 1/VC1/L/Chua-1-VC1-L-chunk-0145
|   |   |-- 1/VC1/U/Chua-1-VC1-U-chunk-0070
|   |   |-- 1/VC2/L/Chua-1-VC2-L-chunk-0067
|   |   |-- 1/VC2/U/Chua-1-VC2-U-chunk-0141
|   |   |-- 2/IL/L/Chua-2-IL-L-chunk-0039
|   |   |-- 2/IL/U/Chua-2-IL-U-chunk-0051
|   |   |-- 2/VC1/L/Chua-2-VC1-L-chunk-0055
|   |   |-- 2/VC1/U/Chua-2-VC1-U-chunk-0052
|   |   |-- 2/VC2/L/Chua-2-VC2-L-chunk-0058
|   |   `-- 2/VC2/U/Chua-2-VC2-U-chunk-0132
|   |-- Lyapunov/
|   |   `-- Lyapunov1a-chunk-0005
|   |-- Nichols-Plot/
|   |   `-- Inverted-Pendulum/Nichols-Plot-Inverted-Pendulum-Fails-1-6-chunk-0029
|   |-- RL-high-pass-circuit/
|   |   `-- RL-high-pass-circuit-gain-chunk-0005
|   |-- asin/
|   |   |-- 8/asin-8-chunk-0044
|   |   `-- 8/vars4/asin-8-vars4-chunk-0050
|   |-- atan/
|   |   |-- problem/1/atan-problem-1-chunk-0069
|   |   |-- problem/1/weak/atan-problem-1-weak-chunk-0040
|   |   |-- problem/2/atan-problem-2-chunk-0039
|   |   |-- problem/2/sqrt/atan-problem-2-sqrt-chunk-0041
|   |   |-- problem/2/sqrt/weak/atan-problem-2-sqrt-weak-chunk-0012
|   |   |-- problem/2/weak/atan-problem-2-weak-chunk-0063
|   |   |-- problem/2/weak2/atan-problem-2-weak2-chunk-0073
|   |   |-- problem/2/weakT/atan-problem-2-weakT-chunk-0049
|   |   |-- vega/3/atan-vega-3-chunk-0416
|   |   `-- vega/3/weak/atan-vega-3-weak-chunk-0604
|   |-- bottom-plate-mixer/
|   |   `-- bottom-plate-mixer-chunk-0047
|   |-- cbrt/
|   |   |-- 3/cbrt-problem-3-chunk-0068
|   |   `-- 3/weak/cbrt-problem-3-weak-chunk-0063
|   |-- cos/
|   |   |-- 3410/cos-3410-b-chunk-0020
|   |   |-- problem/12/cos-problem-12-chunk-0007
|   |   `-- problem/5/cos-problem-5-chunk-0044
|   |-- exp/
|   |   |-- 361/exp-361-neg-6-e-chunk-0216
|   |   |-- 361/6/exp-361-6-weak-chunk-0804
|   |   |-- problem/10/2/exp-problem-10-2-chunk-0131
|   |   |-- problem/10/3/exp-problem-10-3-chunk-0086
|   |   `-- problem/10/3/weak/exp-problem-10-3-weak-chunk-0036
|   |-- log/
|   |   `-- fun-ineq/log-fun-ineq-g-chunk-0036
|   |-- polypaver/
|   |   |-- bench-exp-3d/polypaver-bench-exp-3d-chunk-0068
|   |   |-- bench-sqrt-3d/polypaver-bench-sqrt-3d-chunk-0207
|   |   |-- sqrt-circles/1a/polypaver-sqrt-circles-1a-chunk-0011
|   |   |-- sqrt-circles/2a/polypaver-sqrt-circles-2a-chunk-0017
|   |   |-- sqrt43-int/polypaver-sqrt43-int-chunk-0050
|   |   |-- sqrt43-int/3vars/polypaver-sqrt43-int-3vars-chunk-0027
|   |   `-- sqrt43-int/4vars/polypaver-sqrt43-int-4vars-chunk-0031
|   |-- sin/
|   |   |-- 344/3/sin-344-3-chunk-0020
|   |   |-- 344/4/sin-344-4-chunk-0045
|   |   |-- cos/sin-cos-346-b-chunk-0188
|   |   |-- problem/10/sin-problem-10-2-chunk-0027
|   |   |-- problem/7/sin-problem-7-chunk-0339
|   |   |-- problem/7/weak/sin-problem-7-weak-chunk-0134
|   |   |-- problem/7/weak2/sin-problem-7-weak2-chunk-0110
|   |   |-- problem/8/sin-problem-8-chunk-0039
|   |   `-- problem/8/weak/sin-problem-8-weak-chunk-0035
|   `-- sqrt/
|       |-- 1mcosq/7/sqrt-1mcosq-7-chunk-0225
|       |-- 1mcosq/8/sqrt-1mcosq-8-chunk-0208
|       |-- circles/2/sqrt-circles-2-chunk-0011
|       |-- circles/3/sqrt-circles-3-chunk-0005
|       |-- problem/sqrt-problem-Melquiond2-chunk-0095
|       |-- problem/12vars3/sqrt-problem-12vars3-chunk-0104
|       |-- problem/13/sqrt-problem-13-chunk-0024
|       |-- problem/13/sqrt1/sqrt-problem-13-sqrt1-chunk-0042
|       `-- problem/13/vars5/sqrt-problem-13-vars5-chunk-0027
`-- zankl/
    `-- matrix-5-all-19
\end{lstlisting}

\clearpage
\section{Pairwise Runtime Comparisons}
\label{app:helms-sota-scatter}

\begin{figure}[htbp]
\centering
\begin{subfigure}[t]{0.48\textwidth}
\centering
\includegraphics[width=\linewidth]{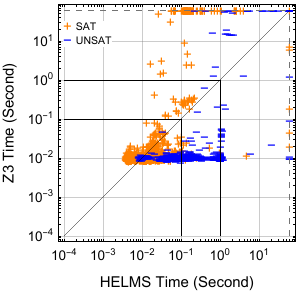}
\caption{Z3}\label{fig:app-helms-z3}
\end{subfigure}\hfill
\begin{subfigure}[t]{0.48\textwidth}
\centering
\includegraphics[width=\linewidth]{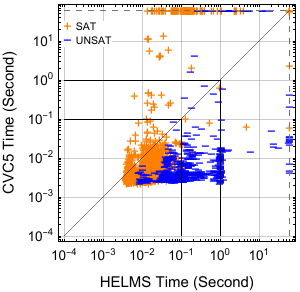}
\caption{CVC5}\label{fig:app-helms-cvc5}
\end{subfigure}
\par\medskip
\begin{subfigure}[t]{0.48\textwidth}
\centering
\includegraphics[width=\linewidth]{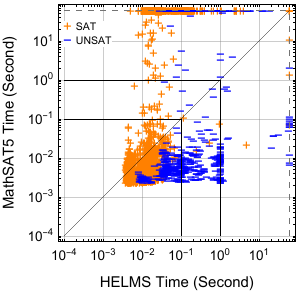}
\caption{MathSAT5}\label{fig:app-helms-mathsat5}
\end{subfigure}\hfill
\begin{subfigure}[t]{0.48\textwidth}
\centering
\includegraphics[width=\linewidth]{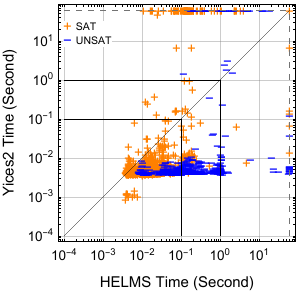}
\caption{Yices2}\label{fig:app-helms-yices2}
\end{subfigure}
\caption{Recorded runtimes on all $2260$ instances, with HELMS on the horizontal
axis. Orange $+$ and blue $-$ denote SAT and UNSAT.}
\label{fig:app-helms-sota-scatter}
\end{figure}

\end{document}